\documentclass[lettersize,journal]{IEEEtran}
\usepackage{amsmath,amsfonts}
\usepackage{algorithmic}
\usepackage{algorithm}
\usepackage{array}
\usepackage[caption=false,font=normalsize,labelfont=sf,textfont=sf]{subfig}
\usepackage{textcomp}
\usepackage{stfloats}
\usepackage{url}
\usepackage{verbatim}
\usepackage{graphicx}
\usepackage{cite}
\usepackage{booktabs}
\usepackage{multirow}
\usepackage{makecell}
\usepackage{adjustbox}
\usepackage{xcolor}
\usepackage[normalem]{ulem}
\usepackage{tabularx}
\usepackage{subcaption}
\usepackage{caption}
\usepackage{csquotes}
\usepackage{titlesec}
\usepackage{microtype}
\titlespacing*{\section}{0pt}{1.2ex plus 1ex minus .2ex}{0.8ex plus .2ex}
\titlespacing*{\subsection}{0pt}{0.8ex plus 1ex minus .2ex}{0.5ex plus .2ex}

\begin{document}

\title{From User Recognition to Activity Counting: An Identity-Agnostic Approach to Multi-User WiFi Sensing}
\author{Kemal Bayik, Olayinka Ajayi, Daniel Roggen, Philip Birch
\thanks{K. Bayik, O. Ajayi, D. Roggen, P. Birch are with the School of Engineering
and Informatics, University of Sussex, Brighton, East Sussex, UK
(e-mail: K.Bayik@sussex.ac.uk, Olayinka.Ajayi@sussex.ac.uk, D.Roggen@sussex.ac.uk, P.M.Birch@sussex.ac.uk).}%
}

\maketitle

\begin{abstract}
WiFi Channel State Information (CSI) enables device-free human activity
recognition, but multi-user methods typically assume a fixed set of known
users and degrade on unseen individuals. We reformulate multi-user
recognition as identity-agnostic activity counting, estimating how many
users perform each activity without assigning actions to individuals. For a
controlled comparison, identity-dependent predictions are converted to
activity counts and evaluated with the same integer-valued metrics using a
shared pipeline that differs only in output head and objective. On WiMANS,
identity-agnostic counting achieves lower integer-count MAE than derived
identity-dependent counts (0.0815 versus 0.0909 on a 0--5 count scale)
despite using coarser supervision. This ordering holds across all eight
environment--band conditions, while derived counts lead only in exact-match
accuracy. Under leave-users-out evaluation, identity-dependent macro-F1
drops from 80.02\% to 33.85\%, its derived counts perform only slightly better 
than an all-zero predictor, and the exact-match advantage reverses, while
identity-agnostic counting leads on all counting metrics. Silhouette analysis
shows that identity-dependent embeddings cluster by user identity, whereas
identity-agnostic embeddings cluster by activity composition. Activity
counting therefore provides a more accurate and robust formulation for
scene-level activity composition estimation without requiring a predefined
user set.
\end{abstract}


\section{Introduction}

Recent advances in wireless sensing have extended the capabilities of 
commodity WiFi infrastructures beyond traditional communication, 
transforming them into a pervasive and device-free modality for 
perceiving the physical world
\cite{sai2026machine,tan2022Commodity, chen2022}. In particular, 
Channel State Information (CSI), which characterizes the fine-grained 
propagation properties of wireless signals across subcarriers and 
antennas, provides rich information about the interaction between radio 
signals and human motion. Compared with traditional vision-based or 
wearable sensing approaches, CSI-based systems offer several advantages 
such as non-intrusive sensing, reduced privacy concerns, and the 
ability to operate using widely deployed commodity infrastructures. 
These properties have motivated extensive research efforts toward 
leveraging CSI for a broad range of human-centric sensing tasks. 
Consequently, CSI has demonstrated promising performance in diverse 
applications such as device-free activity recognition 
\cite{WiAct,DeepSense}, gesture recognition 
\cite{GMM-Gesture,FingerPass}, fall detection 
\cite{RT-Fall,FallDetectionSpectogram}, respiration monitoring 
\cite{MultiSense}, and intrusion detection 
\cite{PetFree,Human-NonHuman}. Despite these significant advancements, 
many CSI-based sensing systems are primarily designed for single-user 
scenarios under controlled environments, while real-world deployments 
often involve multiple users interacting simultaneously within dynamic 
indoor environments.

Recognizing human activities from WiFi signals becomes considerably
more difficult in multi-user scenarios, as concurrent body movements
cause signal distortion, interference, and overlapping reflections
in the wireless channel~\cite{whyMultiUserHard}. To address this
complexity, existing studies adopt identity-dependent formulations
that predict activities separately for each predefined
subject~\cite{WISDOM,MultiSenseX,InceptionTime}. These methods
assume a fixed set of known users and frame recognition as a
user-indexed prediction task. While effective under controlled
conditions, such formulations limit practical deployment because
real-world environments often include unseen users and a changing
number of subjects. Current systems are typically evaluated using
closed-set protocols where all subjects appear during training,
raising uncertainty as to whether reported performance reflects
genuine activity understanding or partially relies on user-specific
signal characteristics. This uncertainty reflects a cost that is
intrinsic to the formulation. Requiring a model to decide which user
performs which activity forces it to encode user-specific signal
characteristics, which are informative for enrolled individuals and
misleading for anyone else. The requirement is also stronger than
most deployments need, as these ask only what is happening in a
scene rather than who is doing it.

In this work we remove that requirement. Instead of assigning
activities to predefined individuals, we propose an
identity-agnostic formulation that we refer to as activity
counting: estimating the number of occurrences of each activity
type at a given time, without associating actions with specific
individuals. This shifts the task from user-indexed classification
to scene-level activity composition estimation and removes the
need for fixed user sets. We develop and evaluate both
formulations on the WiMANS dataset \cite{WiMANS}, implemented on a
common pipeline built around a standard convolutional backbone, so
that any difference between them follows from the output structure
and training objective rather than from architectural asymmetry.
Because the two produce outputs of different kinds, user-wise predictions are
converted into derived activity counts, which places them on
identical metrics and makes the comparison possible at all. As the 
identity-dependent side of this pipeline substantially exceeds every
baseline evaluated here, the comparison is between two strong systems rather
than against a weakened control. The main contributions are as follows:

\begin{itemize}

\item We propose an identity-agnostic formulation for multi-user
wireless activity analysis that replaces user-indexed
prediction with scene-level activity counting, removing the
need for a predefined set of users.

\item Using matched count metrics, we show that identity-agnostic counting is 
more accurate and robust than counts derived from identity-dependent predictions 
across standard, unseen-user, and unseen-environment settings. Feature-space analyses 
reveal representations organized around activity composition rather than user identity, 
while linear probes on frozen features show that the counting advantage persists under an identical readout.

\item We show that ImageNet-pretrained convolutional features transfer
effectively to multi-user CSI projections, where four of the five CSI
architectures fitted with the same counting head yield negative
$R^2_{\mathrm{int}}$. Permutation and from-scratch ablations suggest that performance depends
primarily on pretrained filters and preserved temporal locality rather than
precise projection geometry.

\end{itemize}

\section{Related Works}

\subsection{Single-User Activity Recognition}

Early CSI-based recognition systems relied on handcrafted signal features 
\cite{WiAct, WiHAR}, but deep learning approaches have largely replaced 
manual feature engineering. Architectures ranging from autoencoder-enhanced 
convolutional-recurrent networks \cite{DeepSense}, hybrid CNN-GRUs 
\cite{CNN-GRU}, attention-based sequence models \cite{ABLSTM}, and 
transformers \cite{THAT, ViT} to pure 2D-CNN designs \cite{grayscaleCSI} 
have achieved high performance by capturing complex spatio-temporal patterns 
in the wireless channel. However, they are fundamentally formulated for single-user 
settings, whereas real-world environments often involve multiple people 
performing different activities simultaneously.

\subsection{Multi-User Activity Recognition}

A common strategy in early multi-user WiFi sensing was to first separate 
user-specific signal components and then apply per-user activity classifiers. WISDOM \cite{WISDOM}, 
MultiTrack \cite{MultiTrack}, and IMar \cite{IMar} follow this general strategy using, respectively, 
signal sorting, multi-link profile reconstruction, and tensor decomposition to isolate user-specific 
information, while WiMU \cite{WiMU} instead generates virtual multi-user samples from single-user 
training data. Collectively, these systems demonstrate that multi-user recognition is feasible for 
several simultaneous users, but they are all evaluated on private datasets, which limits 
reproducibility and makes systematic comparison difficult.

More recent work has shifted toward publicly available benchmarks, with WiMANS \cite{WiMANS} 
emerging as a common evaluation platform. On this dataset, MultiSenseX \cite{MultiSenseX} studies 
joint localization and activity recognition in multi-occupant settings using a multi-label 
transformer design, Wang et al. \cite{InceptionTime} combine inception modules with 
attention mechanisms to capture both short-term and long-term patterns in CSI, and 
WiMAR \cite{WiMAR} emphasizes the separation of dynamic 
and static signal components and processes the motion-related component using a CNN--GRU--attention 
pipeline. More recently, TSDBN~\cite{TSDBN} proposes a lightweight dual-branch design
that processes a temporally compressed CSI representation through a
convolution-augmented attention branch and a parallel 2D convolutional
branch, and reports identity, activity, and location recognition on WiMANS
under both bands. These studies provide a more comparable basis for evaluation, but they also share 
an important limitation, as performance is reported primarily in terms of accuracy.
This is problematic for WiMANS because the dataset exhibits a strong class imbalance induced by 
the \textit{null} label (represented as \textit{nan} in the dataset codebase), which denotes the absence of a user in a given slot. Since multiple user 
slots may be unoccupied in the same sample, a model with weak discriminative ability can still 
achieve misleadingly high accuracy by over-predicting \textit{null}. This makes accuracy an 
unreliable indicator of true recognition performance, and we argue that macro-F1 score provides a 
more faithful evaluation. 

More recently, UN-2DCNN \cite{UN2DCNN} 
reports state-of-the-art results, but because some preprocessing details are not 
fully specified, we were unable to reproduce the pipeline reliably and therefore exclude it from 
direct comparison.

Despite their methodological diversity, all of the above approaches share a common assumption: 
activities are predicted separately for each predefined individual, tying model outputs directly to a 
fixed set of known users. Although IMar includes action counting, this counting remains 
user-indexed and still depends on prior knowledge of user count together with successful per-user 
signal decomposition. A recent exception is AMAR~\cite{AMAR}, which formulates multi-user HAR as
set prediction: the predictions of a set of query slots are matched to
ground-truth activities by the Hungarian algorithm, so the output carries no
identity indexing and activity counts can be obtained by aggregating the
slots. Its slot count must nonetheless be fixed in advance, and its evaluation is
confined to environment-specific standard splits, so the effect of removing
identity supervision on cross-user generalization remains an open question.
In contrast, this work reformulates the task as
scene-level activity composition estimation, removing the dependence on user
identity and count altogether. This enables more flexible generalization to
settings in which the number and identity of active users may be unknown at
inference time.

Beyond these specific studies, reproducibility remains a significant concern across the multi-user sensing 
literature. As highlighted by Guarino et al. \cite{reproducibility}, a large portion of published 
works lack essential experimental details, publicly available datasets, or accessible preprocessing 
code, making systematic comparison across studies challenging.

\subsection{Generalization Across Users and Environments}

Generalization across unseen users and environments remains one of the central challenges in 
WiFi sensing. As identified by Wang et al. \cite{generalizability}, three major factors hinder 
generalization: device heterogeneity, human body diversity, and environmental diversity. Even in 
single-user scenarios, differences in body shape, motion style, and room layout can produce 
substantial variation in the measured signal. 

Existing work has mainly studied generalization along two axes: unseen users and unseen 
environments. For cross-user robustness, Widar3.0~\cite{widar3} derives velocity profiles that
capture kinematic characteristics independent of the performer or location. Meta-learning approaches such as MetaFormer~\cite{metaformer} and Wi-Learner 
\cite{wilearner} further improve adaptation to unseen users under limited supervision, while 
CrossID~\cite{crossid} explicitly models identity-related information as a transferable 
representation. FewSense~\cite{FewSense} pre-trains a feature extractor
offline and classifies novel classes in unseen domains by cosine similarity,
fine-tuning on a handful of labelled target samples, while
WiTeacher~\cite{WiTeacher} removes the need for target labels altogether by
combining a Mean Teacher framework with generated target-like samples and
adaptive label smoothing.

On the environment side, AirFi~\cite{airfi} proposes a domain generalization 
framework that extracts and augments common features across multiple training 
environments to enable recognition 
in unseen environments without target data. WiHARAN~\cite{wiharan} aligns 
the joint distribution of features and labels across 
environments through adversarial learning, while 
DA-HAR~\cite{dahar} extracts environment-independent features using a dual 
adversarial network with pseudo-label prediction. Additionally, 
Zhang et al.~\cite{privacy} apply the Johnson-Lindenstrauss transform to 
generate differentially private representations for cross-environment sharing. 
What these methods have in common is that they counter domain shift
with an additional resource: labelled or unlabelled samples from the
target domain, data from several training environments, or
hand-designed domain-invariant features that require particular
hardware configurations. They are also all developed for single-user
settings. The present work instead asks whether the choice of problem
formulation itself affects sensitivity to unseen users, without any of
these and in the multi-user setting.

\begin{figure*}[t]
\centering

\includegraphics[width=\textwidth]{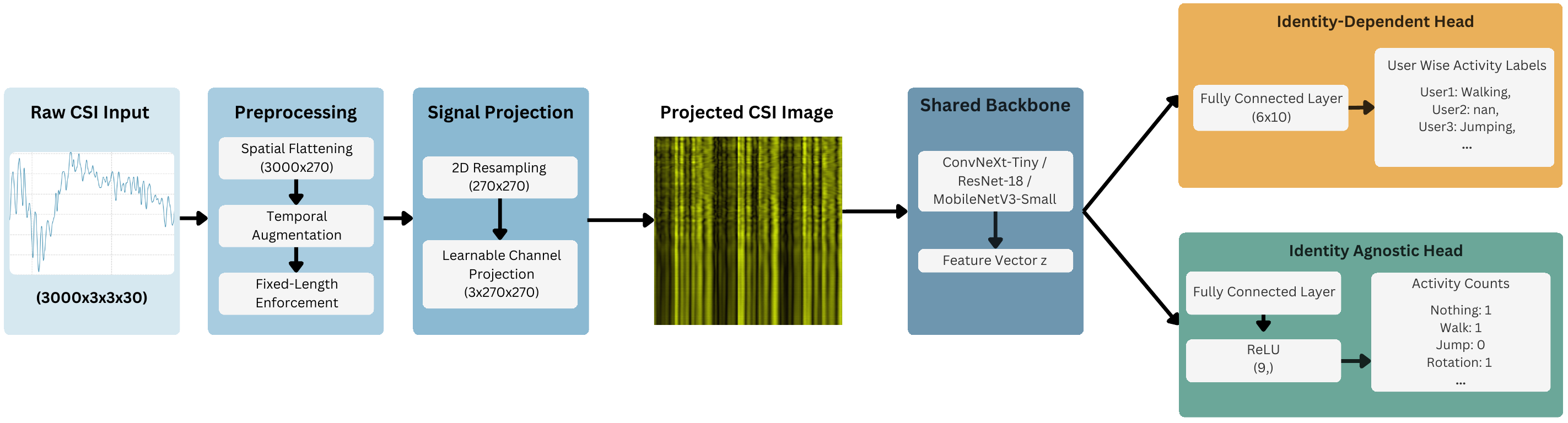}
\caption{Overview of the proposed multi-user CSI sensing framework.
Raw CSI signals are preprocessed and projected into a 2D representation,
processed by a shared pretrained backbone, and fed into either
identity-dependent or identity-agnostic prediction heads.}
\label{fig:architecture}

\end{figure*}

\section{Dataset Description}
\label{sec:dataset}
All experiments were conducted on the publicly available WiMANS dataset
\cite{WiMANS}, which is the first benchmark specifically designed for
WiFi-based multi-user activity sensing. Unlike prior CSI datasets that
contain only a single subject per sample, WiMANS captures simultaneous
activities of multiple users within the same time window. It contains
11,286 CSI samples collected in three indoor environments:
\textit{classroom}, \textit{meeting room}, and \textit{empty room}. Each
scene was recorded on both the 2.4 GHz and 5 GHz bands, so a single band
provides 5,643 samples. Each sample has a duration of 3 seconds and includes 
between 0 and 5 simultaneously active users, drawn from a pool of six participants
(anonymized as User 1--User 6). CSI is recorded using 3 transmit and 3 receive 
antennas over 30 subcarriers at a nominal rate of 1000 packets per second. 
Recordings vary slightly in length because of packet loss and are
reflection-padded to 3000 packets where necessary, yielding an input
dimension of $3000 \times 3 \times 3 \times 30$.

The dataset defines nine daily activities: \textit{Nothing},
\textit{Walking}, \textit{Rotation}, \textit{Jumping}, \textit{Waving},
\textit{Lying Down}, \textit{Picking Up}, \textit{Sitting Down}, and
\textit{Standing Up}. These range from sustained whole-body motion to
brief localised gestures and to \textit{Nothing}, which denotes a user
present in the scene but stationary. Each sample is annotated with
environment label, number of users, user identities, locations, and
per-user activity labels. Activity labels are provided for six identity slots, with absent
users marked as \texttt{nan} in the dataset codebase. We encode
these as a tenth label, $\emptyset$, distinct from the \emph{Nothing}
activity above. The samples are distributed equally across the three
environments.

Three properties of WiMANS are important for this work. First, multiple
users may perform different activities simultaneously, creating overlapping
signal reflections that make identity-dependent recognition challenging.
Second, occupancy varies across samples and is balanced by design: the
dataset contains equal numbers of samples for two, three, four, and five
active users (15.79\% each), twice as many single-user samples (31.58\%), and a
smaller set of empty scenes (5.26\%), giving a mean of 2.53 active users. As
these users are distributed over nine activity classes, most entries of the
ground-truth count vector are zero. Third, the six-participant pool allows
the users seen during training to be disjoint from those seen at test time,
which is what the leave-users-out protocol of Section~\ref{sec:metrics}
exercises.

\section{Methodology}

\subsection{Problem Formulation}

Let $X \in \mathbb{R}^{T \times 3 \times 3 \times 30}$ denote a CSI sample
with $T=3000$ temporal packets. Two formulations are considered: 
Identity-Dependent Activity Recognition (ID) and Identity-Agnostic 
Activity Counting (IA).
In the identity-dependent activity recognition formulation, the task is defined as
predicting the activity of each predefined user slot separately.
Let $U=6$ denote the number of user slots defined by the dataset, of which
at most five are occupied in any scene. For each
user $u \in \{1,\dots,U\}$, the activity label is defined as
$y_u \in \mathcal{A}_{\text{dep}} = \{\emptyset, a_1, a_2, \dots, a_9\}$. 
Writing $\mathbf{y}_u \in \{0,1\}^{10}$ for the one-hot encoding of $y_u$,
the target for a given CSI sample is the matrix $Y \in \{0,1\}^{U \times 10}$
whose $u$-th row is $\mathbf{y}_u$.
The model learns a mapping $f_\theta : X \rightarrow \hat{Y}$, where
$\hat{Y} \in \mathbb{R}^{U \times 10}$ contains class scores for each user
slot; the predicted label for slot $u$ is $\arg\max_j \hat{Y}_{u,j}$.
The output is tied to a fixed user roster and therefore assumes consistent
identity indexing in closed-set operation; the leave-users-out protocol
deliberately violates this assumption.

In the identity-agnostic formulation, the task is to estimate the number of
occurrences of each activity type within the scene, without associating
actions with specific users. For a given CSI sample, the target output is
$\mathbf{c} = [c_1, \dots, c_9] \in \mathbb{N}_0^{9}$, where 
each element $c_k$ represents the number of users performing
activity $a_k$ in that sample. The ground-truth count for class
$a_k$ is computed as $c_k = \sum_{u=1}^{U} \mathbb{I}[y_u = a_k]$.
For example, two users walking and one sitting down gives
$c_{\text{walk}}=2$, $c_{\text{sit\_down}}=1$, and zero elsewhere. The
model learns a mapping $g_\theta : X \rightarrow \hat{\mathbf{c}}$,
where $\hat{\mathbf{c}} \in \mathbb{R}_{\ge 0}^{9}$ are continuous
non-negative predictions obtained through a regression head. For
discrete counting evaluation, predictions are rounded element-wise
to the nearest integer.

\subsection{CSI Representation, Projection, and Backbone}
\label{sec:projection}

CSI amplitude measurements were used as the input representation. As the first 
preprocessing step, the spatial and frequency 
dimensions ($3 \times 3$ antenna pairs and $30$ subcarriers) were 
flattened to obtain a time-feature matrix $X_f \in \mathbb{R}^{T \times 270}$, 
preserving temporal ordering. In this representation, the rows preserve the temporal packet sequence, while
the columns encode the antenna-pair and subcarrier responses. Two-dimensional
convolutional filters can therefore capture local temporal variations while
aggregating information across the wireless channels. During training, temporal warping was 
applied with probability 0.5 by resampling the temporal axis using a scaling 
factor drawn from $[0.95, 1.05]$. All samples were then resized to a fixed 
length of 3000 packets by truncation or reflection padding.
To transfer ImageNet-pretrained vision backbones to CSI sensing, the 
resulting matrix was interpreted as a single-channel 2D signal and resized 
with bicubic interpolation to $X_r \in \mathbb{R}^{270 \times 270}$. This 
resolution preserves the flattened feature dimension without additional 
compression. A learnable $1 \times 1$ convolution then projected $X_r$ to 
$X_{rgb} \in \mathbb{R}^{3 \times 270 \times 270}$, producing a three-channel 
representation compatible with the pretrained convolutional networks.
This projection does not assume semantic equivalence between CSI matrices and
natural images; rather, it provides a regular two-dimensional structure that
can be adapted by pretrained convolutional filters during end-to-end
fine-tuning. Section~\ref{sec:ablation} examines which of its two axes the
resulting model actually relies on.

The projected tensor $X_{rgb}$ is processed by an ImageNet-1K-pretrained
convolutional backbone. Three architectures with different capacity levels
were evaluated: ConvNeXt-Tiny~\cite{convnext}, ResNet-18~\cite{resnet}, and
MobileNetV3-Small~\cite{mobilenet}. For each model, the classification layer
was removed while retaining the convolutional feature extractor and global
average pooling, yielding a feature vector $\mathbf{z} \in \mathbb{R}^{d}$.
This shared representation was then passed to task-specific prediction
heads. All backbones were fine-tuned end-to-end. The overall architecture is
illustrated in Fig.~\ref{fig:architecture}.

\subsection{Task-Specific Prediction Heads and Objectives}

Given the shared feature representation $\mathbf{z} \in \mathbb{R}^{d}$
extracted by the backbone, the final prediction layer and training objective
differ depending on the task formulation. The source code will be released
upon acceptance.

\textbf{Identity-dependent recognition:} The objective was to predict the
activity label for each of the $U=6$ predefined user slots across $10$
classes (including the $\emptyset$ class). A fully connected layer mapped
the feature vector to class logits:
$\mathbf{z} \rightarrow \hat{Y} \in \mathbb{R}^{U \times 10}$. To mitigate
the class imbalance caused by the frequent $\emptyset$ class, the model was
optimized using a focal loss~\cite{focalloss} ($\gamma = 2.0$) applied
independently to each user slot. Under the leave-users-out
protocol the head has three slots rather than six.

\textbf{Identity-agnostic counting:} A single fully connected layer mapped
the features to continuous count predictions for the $9$ distinct
activities (excluding $\emptyset$):
$\mathbf{z} \rightarrow \hat{\mathbf{c}} \in \mathbb{R}^{9}$. A ReLU
activation ensured non-negative outputs, and the model was trained with a
Mean Squared Error (MSE) objective between predicted and ground-truth
activity counts.

\begin{table*}[!t]
\centering
\caption{Performance and complexity under the Standard protocol
(5~GHz, all environments). Complexity measurements correspond to the
identity-agnostic configurations. Bold denotes the best value per column.
CPU latency is measured per sample using FP32, batch size 1, and a single thread.}
\label{tab:baseline_complexity}

\scriptsize
\setlength{\tabcolsep}{2.0pt}
\renewcommand{\arraystretch}{1.08}

\begin{adjustbox}{max width=\textwidth}
\begin{tabular}{lcccccccccccc}
\toprule
& \multicolumn{1}{c}{ID Recognition}
& \multicolumn{4}{c}{ID-Derived Counts}
& \multicolumn{4}{c}{IA Counting}
& \multicolumn{3}{c}{Complexity} \\
\cmidrule(lr){2-2}
\cmidrule(lr){3-6}
\cmidrule(lr){7-10}
\cmidrule(lr){11-13}

Model
& \makecell{Macro-F1\\(\%)}
& \makecell{Cell Acc.\\(\%)}
& \makecell{Exact Match\\(\%)}
& $\mathrm{MAE}_{\mathrm{int}}$
& $R^2_{\mathrm{int}}$
& \makecell{Cell Acc.\\(\%)}
& \makecell{Exact Match\\(\%)}
& $\mathrm{MAE}_{\mathrm{int}}$
& $R^2_{\mathrm{int}}$
& \makecell{Params\\(M)}
& \makecell{MACs\\(G)}
& \makecell{Latency\\(ms)} \\
\midrule

2D-CNN~\cite{grayscaleCSI}
& 26.47 $\pm$ 0.68 & 70.54 $\pm$ 0.07 & 11.43 $\pm$ 0.78
& 0.3693 $\pm$ 0.0049 & $-0.3895 \pm 0.0175$
& 74.03 $\pm$ 0.38 & 6.11 $\pm$ 0.62
& 0.3028 $\pm$ 0.0066 & $-0.0316 \pm 0.0412$
& 0.8874 & 0.8421 & 15.1 \\

CNN-LSTM~\cite{CLSTM}
& 25.83 $\pm$ 0.75 & 69.51 $\pm$ 0.09 & 10.83 $\pm$ 1.03
& 0.3876 $\pm$ 0.0057 & $-0.4838 \pm 0.0277$
& 76.19 $\pm$ 0.16 & 5.46 $\pm$ 1.17
& 0.2923 $\pm$ 0.0199 & $-0.0991 \pm 0.1563$
& 5.3682 & 0.8587 & 32.1 \\

ABLSTM~\cite{ABLSTM}
& 27.68 $\pm$ 0.78 & 71.17 $\pm$ 0.51 & 11.54 $\pm$ 1.34
& 0.3588 $\pm$ 0.0082 & $-0.3327 \pm 0.0196$
& 77.22 $\pm$ 0.52 & 5.67 $\pm$ 1.08
& 0.2780 $\pm$ 0.0088 & $-0.0272 \pm 0.1010$
& 4.2706 & 0.3936 & 157.4 \\

THAT~\cite{THAT}
& 41.28 $\pm$ 1.48 & 77.85 $\pm$ 0.22 & 24.95 $\pm$ 0.78
& 0.2691 $\pm$ 0.0052 & 0.0447 $\pm$ 0.0082
& 78.35 $\pm$ 0.42 & 4.99 $\pm$ 0.50
& 0.2849 $\pm$ 0.0002 & $-0.2002 \pm 0.0099$
& 4.8885 & 0.5492 & 24.6 \\

TSDBN~\cite{TSDBN}
& 37.38 $\pm$ 1.38 & 74.78 $\pm$ 0.52 & 23.98 $\pm$ 1.88
& 0.3114 $\pm$ 0.0095 & $-0.1637 \pm 0.0531$
& 77.16 $\pm$ 0.57 & 17.18 $\pm$ 1.25
& 0.2643 $\pm$ 0.0027 & 0.1219 $\pm$ 0.0250
& 1.3612 & 0.1125 & \textbf{7.1} \\

AMAR w/o RVQ~\cite{AMAR}
& -- & -- & -- & -- & --
& 83.76 $\pm$ 0.34 & 44.73 $\pm$ 1.68
& 0.1824 $\pm$ 0.0030 & 0.4244 $\pm$ 0.0286
& \textbf{0.3117} & 0.1947 & 14.6 \\

\midrule

Ours (MobileNetV3-Small)
& 65.97 $\pm$ 0.55 & 85.80 $\pm$ 0.34 & 47.80 $\pm$ 1.33
& 0.1601 $\pm$ 0.0036 & 0.5008 $\pm$ 0.0221
& 86.69 $\pm$ 0.22 & 37.11 $\pm$ 0.76
& 0.1425 $\pm$ 0.0035 & 0.5947 $\pm$ 0.0100
& 0.9322 & \textbf{0.0840} & 9.4 \\

Ours (ResNet-18)
& 66.14 $\pm$ 0.21 & 86.21 $\pm$ 0.14 & 51.37 $\pm$ 0.85
& 0.1552 $\pm$ 0.0025 & 0.5174 $\pm$ 0.0230
& 87.05 $\pm$ 0.83 & 37.23 $\pm$ 4.38
& 0.1436 $\pm$ 0.0113 & 0.5514 $\pm$ 0.0632
& 11.1811 & 2.7454 & 57.8 \\

Ours (ConvNeXt-Tiny)
& \textbf{80.02 $\pm$ 0.31} & \textbf{91.71 $\pm$ 0.19}
& \textbf{68.50 $\pm$ 0.49} & \textbf{0.0909 $\pm$ 0.0028}
& \textbf{0.7308 $\pm$ 0.0058}
& \textbf{92.25 $\pm$ 0.11} & \textbf{57.75 $\pm$ 0.85}
& \textbf{0.0815 $\pm$ 0.0018} & \textbf{0.7769 $\pm$ 0.0111}
& 27.8255 & 5.9958 & 117.0 \\

\bottomrule
\end{tabular}
\end{adjustbox}

\end{table*}

\subsection{Implementation and Evaluation Details}
\label{sec:metrics}
The models were trained on an NVIDIA 
RTX A4000 GPU for 50 epochs using AdamW (weight decay $10^{-2}$), a batch 
size of 16, and gradient clipping (max norm 1.0). A two-stage 
learning rate schedule was applied: a linear warm-up during the first 10\% of epochs 
followed by cosine annealing. The initial learning rate was set to $10^{-3}$ 
for the projection module and $10^{-4}$ for the backbone and prediction heads.
Two baselines used paper-specific training schedules rather than the common 50-epoch setup: 
TSDBN for 200 epochs under the same optimiser, and AMAR for 300 epochs with Adam following 
its authors' configuration. The remaining baselines follow the configuration above.

\textbf{Evaluation Metrics:}
Identity-dependent recognition was evaluated with macro-F1, which is
preferred to overall accuracy because of the class imbalance induced by the
frequent $\emptyset$ label. For scene-level comparison, the user-wise
predictions were converted into derived counts. Let
$\hat{y}_u=\arg\max_j\hat{Y}_{u,j}$ denote the predicted class for user
slot $u$; the derived count for activity $a_k$ was computed as
$\hat{c}^{\mathrm{dep}}_k=\sum_{u=1}^{U}\mathbb{I}[\hat{y}_u=a_k]$.
Slots predicted as $\emptyset$ therefore contributed to none of the nine
counts, whereas \emph{Nothing} is one of them.
The same conversion was applied to all
identity-dependent models, including the re-implemented baselines.

Identity-agnostic counting produces continuous predictions, which were
evaluated using $\mathrm{MAE}_{\mathrm{cont}}$ and
$R^2_{\mathrm{cont}}$. For direct comparison with the integer-valued
derived counts, these predictions were rounded to the nearest integer;
metrics computed from integer-valued predictions are denoted
$\mathrm{MAE}_{\mathrm{int}}$ and $R^2_{\mathrm{int}}$. MAE was
computed over all activity--sample entries of the count matrix, whereas
$R^2$ was computed separately for each of the nine activity dimensions
and averaged uniformly across them. Cell-wise accuracy measures the
proportion of individual activity counts predicted exactly, whereas
exact-match accuracy requires the complete activity-count vector of a
sample to be correct; both were computed from integer-valued
predictions. All conversions were applied only during evaluation and did
not alter either training objective. 

\textbf{Data Splits and Baselines:} Generalization was evaluated across three protocols: (1) \textit{Standard
Split} (training and testing on the same environments/users), (2)
\textit{Leave-Users-Out (LUO)}, where models were trained on three users
and evaluated on three disjoint users, and (3)
\textit{Leave-One-Environment-Out (LOEO)}. Standard and LOEO retain the six global identity slots. LUO instead uses a
three-slot head: the disjoint training and test identities are independently
mapped, in list order, to local slots 1--3. Every output unit 
therefore receives training supervision, including
$\emptyset$ labels; only its associated physical identity changes at test
time, avoiding unused output units as a confound.
The Standard and LOEO partitions have mean test occupancies of 2.56 and 2.53, whereas the LUO
partitions have 1.25, so counting errors are compared between formulations
within a protocol rather than across protocols. A predictor emitting zero
for every activity attains an $\mathrm{MAE}_{\mathrm{int}}$ of 0.2849,
0.1389, and 0.2807 on the Standard, LUO, and LOEO partitions respectively. All
experiments were repeated over three data partitions and are reported as
mean $\pm$ SD.

The evaluated formulations were benchmarked against five representative CSI
sensing architectures: 2D-CNN~\cite{grayscaleCSI}, CNN-LSTM~\cite{CLSTM},
ABLSTM~\cite{ABLSTM}, THAT~\cite{THAT}, and TSDBN~\cite{TSDBN}, together with 
AMAR without RVQ~\cite{AMAR}, the stronger of the two configurations reported 
by its authors, as an additional counting-oriented baseline. All models
received the same input matrix $X_f$ and the same Standard-protocol
partitions, and each then applies the input transformation specified in its
original work. The domain-shift protocols were applied only to the two formulations within
the proposed pipeline. Except for AMAR, all were fitted with the prediction heads and
objectives described above. AMAR retains its own set-prediction objective and augmentations, with activity
counts obtained by aggregating its six query slots at evaluation time.

Computational complexity was evaluated in terms of total parameter count
and multiply--accumulate operations (MACs) for a single forward pass,
together with mean inference latency per sample. Latency was measured on an 
Intel Xeon W-2223 CPU (3.6 GHz) in FP32 at batch size 1 with a single thread, 
as an average over 500 consecutive forward passes following 50 warm-up passes. 
To examine the learned representations, 768-dimensional feature vectors were
extracted from the final pooled layer preceding the prediction heads, using
the single-user test samples of the Standard partitions, since attributing a
sample to a user identity is only well defined when one user is active.
Silhouette scores were computed in this space under two clustering criteria,
grouping samples either by the identity of the active user or by their
activity-count vector, which for single-user samples reduces to the activity
performed, and the same vectors were projected to two dimensions with t-SNE
for visualisation. Identical samples were used for both formulations.

To separate the contribution of the representation from that of the output
head, a linear count probe was fitted on frozen features. For each
formulation the projection module and backbone were frozen after training and
a single linear layer with ReLU was trained for 50 epochs on the same MSE
counting objective and partitions, without augmentation. The two probes 
therefore share an identical readout and differ only in the representation they receive.

Uncertainty in the cross-formulation comparison was assessed with a paired
sample-level bootstrap over 10{,}000 resamples, clustered by sample
identifier, on the per-sample difference between the derived and
identity-agnostic count MAE. Unlike the across-partition SD, this quantifies
test-sample rather than training-split variability.

\section{Experimental Results}

\subsection{Backbone Selection and Comparison with Existing Models}
\label{sec:backbone}

Three ImageNet-pretrained backbones were evaluated under both formulations
(Table~\ref{tab:baseline_complexity}). ConvNeXt-Tiny achieved the strongest 
predictive performance, reaching a macro-F1 of 80.02 and an 
$\mathrm{MAE}_{\mathrm{int}}$ of 0.0815, while no other backbone reached 67 
in macro-F1 or fell below 0.1425 in $\mathrm{MAE}_{\mathrm{int}}$. It was 
therefore used for all subsequent experiments.

The five identity-dependent baselines reached macro-F1 scores between 25.83
and 41.28; the proposed model exceeded the highest of these, THAT, by 38.7
points. Their derived counts yielded $R^2_{\mathrm{int}} \leq 0.045$, four
of them negative. Under the identity-agnostic formulation, AMAR w/o RVQ produced the strongest
baseline count estimates ($\mathrm{MAE}_{\mathrm{int}} = 0.1824$,
$R^2_{\mathrm{int}} = 0.4244$); the proposed model reduced this error by
55\%. 

TSDBN
followed at 0.2643 with a positive but small $R^2_{\mathrm{int}}$ of 0.1219,
while the remaining four architectures reached
$\mathrm{MAE}_{\mathrm{int}} \geq 0.2780$ with negative $R^2_{\mathrm{int}}$.

Parameter counts and MACs spanned roughly two orders of magnitude across
models, whereas latency varied by a factor of about 20, and the three measures
did not rank the models consistently. AMAR w/o RVQ had the fewest parameters,
MobileNetV3-Small the fewest MACs, and TSDBN the lowest latency at 7.1~ms.
MobileNetV3-Small required 71$\times$ fewer MACs than ConvNeXt-Tiny but was
only 12.4$\times$ faster (9.4 against 117.0~ms per sample). The clearest
discrepancy was ABLSTM, the slowest model overall at 157.4~ms despite
requiring 15$\times$ fewer MACs than ConvNeXt-Tiny, reflecting the sequential
computation of its recurrent layers.

\begin{table*}[!t]
\centering
\caption{Integer-count results across protocols, environments, and bands
(mean $\pm$ SD over three data partitions). Domain-shift experiments use
5~GHz and all environments. Comparisons are within each row; bold denotes
the better counting result. Macro-F1 applies only to ID recognition.}
\label{tab:environment_frequency}
\label{tab:unified_generalization}

\scriptsize
\setlength{\tabcolsep}{1.8pt}
\renewcommand{\arraystretch}{1.08}

\begin{adjustbox}{max width=\textwidth}
\begin{tabular}{llccccccccc}
\toprule
& &
\multicolumn{1}{c}{ID Recognition} &
\multicolumn{4}{c}{ID-Derived Counts} &
\multicolumn{4}{c}{IA Counting} \\
\cmidrule(lr){3-3}
\cmidrule(lr){4-7}
\cmidrule(lr){8-11}

Setting
& Band
& \makecell{Macro-F1\\(\%)}
& \makecell{Cell Acc.\\(\%)}
& \makecell{Exact Match\\(\%)}
& $\mathrm{MAE}_{\mathrm{int}}$
& $R^2_{\mathrm{int}}$
& \makecell{Cell Acc.\\(\%)}
& \makecell{Exact Match\\(\%)}
& $\mathrm{MAE}_{\mathrm{int}}$
& $R^2_{\mathrm{int}}$ \\
\midrule

\multicolumn{11}{l}{\textit{Standard protocol}} \\
\midrule

\multirow{2}{*}{All environments}
& 5~GHz
& 80.02 $\pm$ 0.31
& 91.71 $\pm$ 0.19
& \textbf{68.50 $\pm$ 0.49}
& 0.0909 $\pm$ 0.0028
& 0.7308 $\pm$ 0.0058
& \textbf{92.25 $\pm$ 0.11}
& 57.75 $\pm$ 0.85
& \textbf{0.0815 $\pm$ 0.0018}
& \textbf{0.7769 $\pm$ 0.0111} \\

& 2.4~GHz
& 49.02 $\pm$ 0.35
& 79.83 $\pm$ 0.27
& \textbf{30.97 $\pm$ 1.68}
& 0.2313 $\pm$ 0.0005
& 0.2391 $\pm$ 0.0321
& \textbf{82.25 $\pm$ 0.42}
& 24.68 $\pm$ 1.29
& \textbf{0.1934 $\pm$ 0.0055}
& \textbf{0.4219 $\pm$ 0.0101} \\
\cmidrule(lr){1-11}

\multirow{2}{*}{Classroom}
& 5~GHz
& 71.52 $\pm$ 1.56
& 88.68 $\pm$ 0.86
& \textbf{59.68 $\pm$ 2.43}
& 0.1245 $\pm$ 0.0089
& 0.6053 $\pm$ 0.0378
& \textbf{90.26 $\pm$ 1.15}
& 50.66 $\pm$ 3.68
& \textbf{0.1028 $\pm$ 0.0129}
& \textbf{0.7012 $\pm$ 0.0177} \\

& 2.4~GHz
& 45.10 $\pm$ 1.01
& 79.04 $\pm$ 0.25
& \textbf{25.82 $\pm$ 1.77}
& 0.2409 $\pm$ 0.0010
& 0.1373 $\pm$ 0.0130
& \textbf{82.29 $\pm$ 0.33}
& 21.31 $\pm$ 2.40
& \textbf{0.1914 $\pm$ 0.0033}
& \textbf{0.3890 $\pm$ 0.0369} \\
\cmidrule(lr){1-11}

\multirow{2}{*}{Empty room}
& 5~GHz
& 73.12 $\pm$ 1.49
& 89.38 $\pm$ 0.87
& \textbf{59.59 $\pm$ 4.23}
& 0.1164 $\pm$ 0.0096
& 0.6297 $\pm$ 0.0267
& \textbf{91.35 $\pm$ 0.73}
& 53.58 $\pm$ 4.12
& \textbf{0.0902 $\pm$ 0.0094}
& \textbf{0.7437 $\pm$ 0.0274} \\

& 2.4~GHz
& 41.17 $\pm$ 2.38
& 77.52 $\pm$ 0.39
& \textbf{25.11 $\pm$ 1.26}
& 0.2669 $\pm$ 0.0027
& $-0.0121 \pm 0.0653$
& \textbf{80.03 $\pm$ 0.40}
& 16.98 $\pm$ 0.96
& \textbf{0.2221 $\pm$ 0.0035}
& \textbf{0.2601 $\pm$ 0.0489} \\
\cmidrule(lr){1-11}

\multirow{2}{*}{Meeting room}
& 5~GHz
& 77.22 $\pm$ 0.88
& 89.92 $\pm$ 0.35
& \textbf{62.16 $\pm$ 1.34}
& 0.1102 $\pm$ 0.0026
& 0.6623 $\pm$ 0.0064
& \textbf{91.46 $\pm$ 0.28}
& 54.11 $\pm$ 1.16
& \textbf{0.0899 $\pm$ 0.0022}
& \textbf{0.7421 $\pm$ 0.0121} \\

& 2.4~GHz
& 39.91 $\pm$ 0.91
& 75.77 $\pm$ 0.92
& \textbf{21.04 $\pm$ 1.01}
& 0.2837 $\pm$ 0.0132
& $-0.0120 \pm 0.0546$
& \textbf{79.29 $\pm$ 1.08}
& 15.65 $\pm$ 0.92
& \textbf{0.2288 $\pm$ 0.0088}
& \textbf{0.2650 $\pm$ 0.0026} \\
\midrule

\multicolumn{11}{l}{\textit{Domain shift}} \\
\midrule

Leave-users-out
& 5~GHz
& 33.85 $\pm$ 5.13
& 87.48 $\pm$ 1.19
& 20.68 $\pm$ 2.76
& 0.1335 $\pm$ 0.0137
& $-0.1694 \pm 0.2078$
& \textbf{93.04 $\pm$ 1.62}
& \textbf{54.24 $\pm$ 8.88}
& \textbf{0.0707 $\pm$ 0.0160}
& \textbf{0.4314 $\pm$ 0.1519} \\

Leave-one-environment-out
& 5~GHz
& 38.60 $\pm$ 4.45
& 80.03 $\pm$ 0.91
& 14.78 $\pm$ 0.97
& 0.2469 $\pm$ 0.0075
& 0.0610 $\pm$ 0.0581
& \textbf{84.07 $\pm$ 0.84}
& \textbf{24.24 $\pm$ 3.13}
& \textbf{0.1844 $\pm$ 0.0052}
& \textbf{0.3799 $\pm$ 0.0117} \\

\midrule
\multicolumn{11}{l}{\footnotesize A predictor emitting zero for every activity
attains $\mathrm{MAE}_{\mathrm{int}} = 0.2849$ under the Standard protocol,
$0.1389$ under LUO, and $0.2807$ under LOEO.} \\

\bottomrule
\end{tabular}
\end{adjustbox}
\end{table*}

\subsection{Counting Accuracy under Matched Metrics}
\label{sec:closedset}

Under the Standard protocol the identity-dependent model reached a macro-F1
of $80.02 \pm 0.31$, which serves as the closed-set reference for
Section~\ref{sec:shift}. On matched count metrics (Table~\ref{tab:unified_generalization}),
identity-agnostic counting was more accurate on three of the four counting
metrics: $\mathrm{MAE}_{\mathrm{int}}$ of $0.0815 \pm 0.0018$ against
$0.0909 \pm 0.0028$, cell-wise accuracy of 92.25\% against 91.71\%, and
$R^2_{\mathrm{int}}$ of 0.7769 against 0.7308. The derived counts led on
exact-match accuracy, 68.50\% against 57.75\%, for reasons discussed in 
Section~\ref{sec:metrics_behaviour}. Before rounding, the identity-agnostic predictions
yielded $\mathrm{MAE}_{\mathrm{cont}} = 0.1060 \pm 0.0021$ and
$R^2_{\mathrm{cont}} = 0.8134 \pm 0.0026$. The two formulations also
differed in how their errors were distributed: among samples with at least
one incorrect entry, the derived counts contained 2.4 incorrect entries on
average against 1.7 for identity-agnostic counting. The paired bootstrap yielded 
$\Delta = 0.0095$ in favour of identity-agnostic counting (95\% CI 0.0051--0.0139).

IA counting achieved lower MAE than ID-derived counting with all three
backbones evaluated in our framework. TSDBN showed the same pattern and was
the only identity-dependent baseline whose IA counts attained a positive
$R^2_{\mathrm{int}}$. For the remaining identity-dependent baselines, both
formulations performed close to or below the all-zero predictor, making their
relative ordering less informative. AMAR does not enter this comparison, since 
its set-prediction output has no identity-dependent counterpart. These results 
indicate that the IA advantage is not specific to a particular backbone.

\subsection{Effect of Environment, Frequency Band, and Scene Occupancy}
\label{sec:conditions}

\begin{figure}[t]
\centering
\includegraphics[height=5cm, keepaspectratio]{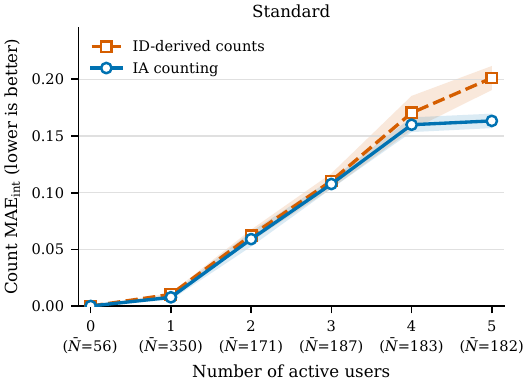}
\caption{Integer-count MAE by active-user count (5~GHz, all environments,
Standard protocol; mean $\pm$ SD over three data partitions). $\bar{N}$
denotes the mean test-sample count per occupancy level.}
\label{fig:user_count}
\end{figure}

Combined training outperformed environment-specific training under both
formulations (Table~\ref{tab:environment_frequency}): macro-F1 reached 80.02
against 71.52 to 77.22, and $\mathrm{MAE}_{\mathrm{int}}$ 0.0815 against
0.0899 to 0.1028 for the identity-agnostic model. Performance was lower at
2.4~GHz in every environment, and the two formulations were affected to
different degrees: $R^2_{\mathrm{int}}$ was negative for the derived counts
in the empty and meeting rooms, whereas the identity-agnostic model retained
0.2601 and 0.2650. The ordering of Section~\ref{sec:closedset} repeated in all eight
environment--band conditions: identity-agnostic counting was more accurate on
every count metric except exact match.
Counting error increased monotonically with
occupancy (Fig.~\ref{fig:user_count}); the two curves remained within 0.011
of each other up to four active users and separated at five, where the
identity-agnostic model reached 0.163 against 0.201.

\subsection{Performance under User and Environment Shift}
\label{sec:shift}

Because the protocol-specific test distributions differ, counting metrics
are compared between formulations within each protocol
(Table~\ref{tab:unified_generalization}).

Under LUO, the macro-F1 of the identity-dependent model decreased from 80.02
to $33.85 \pm 5.13$. On the same test samples, identity-agnostic counting was
better on all four counting metrics, with an $\mathrm{MAE}_{\mathrm{int}}$ of
0.0707 against 0.1335, an exact-match accuracy of 54.24\% against 20.68\%,
and an $R^2_{\mathrm{int}}$ of 0.4314 against $-0.1694$. The exact-match
ordering was thus reversed with respect to the Standard protocol. Before
rounding, the identity-agnostic predictions yielded
$\mathrm{MAE}_{\mathrm{cont}} = 0.0899 \pm 0.0199$ and
$R^2_{\mathrm{cont}} = 0.5199 \pm 0.1160$.

Under LOEO, where the identity-dependent macro-F1 was $38.60 \pm 4.45$, the
identity-agnostic model was again better on all four metrics, with an
$\mathrm{MAE}_{\mathrm{int}}$ of 0.1844 against 0.2469.

Normalizing MAE by each protocol's all-zero-predictor error provides a
distribution-relative comparison across protocols. Under the Standard protocol, IA and
ID-derived MAEs were 0.29 and 0.32 times the zero-predictor error,
respectively. Both ratios increased under domain shift, to 0.51 and 0.96 under
LUO and to 0.66 and 0.88 under LOEO. Thus, under user shift, the ID-derived
MAE (0.1335) nearly equalled the zero-predictor MAE (0.1389), whereas IA
counting produced approximately half that error. The paired bootstrap gave
$\Delta = \mathrm{MAE}_{\mathrm{ID}} - \mathrm{MAE}_{\mathrm{IA}} = 0.0095$
(95\% CI: 0.0051--0.0139) under Standard, increasing to 0.0629
(0.0595--0.0662) under LUO and 0.0625 (0.0584--0.0663) under LOEO, with the
Standard interval disjoint from both.

\begin{table}[t]
\centering
\caption{Silhouette scores in the 768-dimensional embedding space
(5~GHz, all environments, Standard protocol, single-user samples;
mean $\pm$ SD over three data partitions).}
\label{tab:feature_space}
\scriptsize
\setlength{\tabcolsep}{4.0pt}
\renewcommand{\arraystretch}{1.08}
\begin{tabular}{lcc}
\toprule
Clustering criterion & ID Recognition & IA Counting \\
\midrule
User identity        & $0.345 \pm 0.013$ & $-0.037 \pm 0.004$ \\
Activity composition & $0.244 \pm 0.016$ & $0.669 \pm 0.019$ \\
\bottomrule
\end{tabular}
\end{table}

\subsection{Activity-Level Error Structure}
\label{sec:activity}

\begin{figure}[!h]
\centering
\hspace*{-20pt}\includegraphics[width=6cm, keepaspectratio]{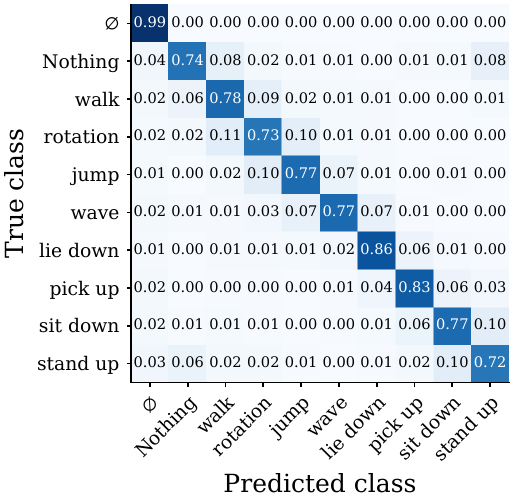}
\caption{Row-normalized identity-dependent confusion matrix averaged over
three data partitions (5~GHz, all environments, Standard protocol).}
\label{fig:confusion}
\end{figure}

The confusion matrix of the identity-dependent model (Fig.~\ref{fig:confusion})
shows near-perfect recovery of $\emptyset$ (0.99) and diagonal
values between 0.72 and 0.86 for the nine activities. Residual mass was
concentrated in approximately symmetric pairs, principally
\textit{sit down}--\textit{stand up} (0.10 in both directions) and
\textit{rotation}--\textit{walk} (0.11 and 0.09); \textit{Nothing} was
additionally confused with \textit{walk} and \textit{stand up}.

\begin{figure}[t]
\centering
\includegraphics[height=4cm, keepaspectratio]{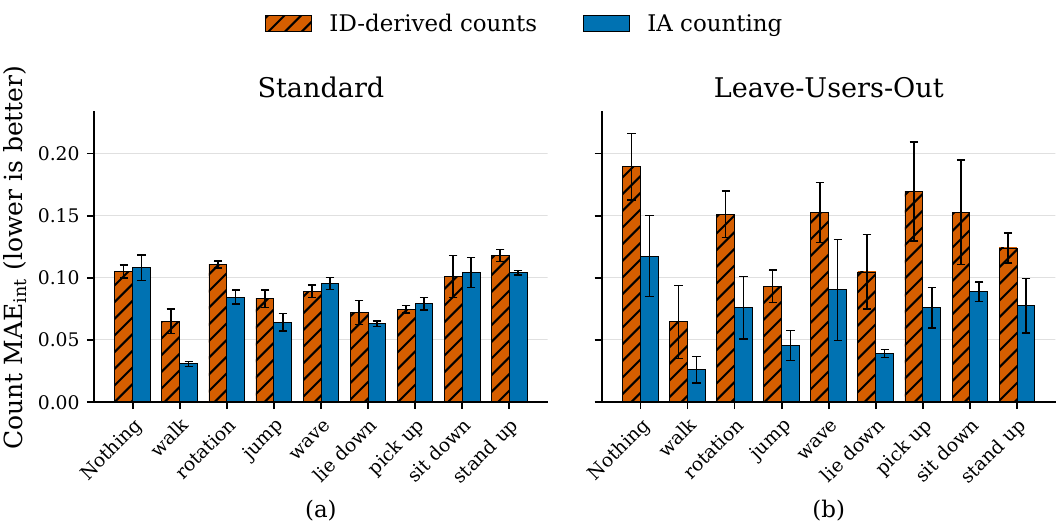}
\caption{Per-activity integer-count MAE for both formulations under the
(a) Standard and (b) Leave-Users-Out protocols (5~GHz, all environments).
Error bars denote SD over three data partitions.}
\label{fig:per_activity}
\end{figure}

Per-activity results provided a finer-grained view of the aggregate pattern
(Fig.~\ref{fig:per_activity}). Under the Standard protocol all nine
activities remained below 0.12 for both formulations, the identity-agnostic
model attaining lower error for five, most clearly \textit{walk} at 0.0307
against 0.0650, and the derived counts attaining lower error for the
remaining four by at most 0.006. Under LUO the identity-agnostic model attained lower error for all
nine, with margins between 0.0384 and 0.0937, and the standard deviations of
the derived counts increased markedly. 

\begin{table}[t]
\centering
\caption{Identity-agnostic ablation results (5~GHz, all environments,
Standard protocol; mean $\pm$ SD over three data partitions).}
\label{tab:ablation}

\scriptsize
\setlength{\tabcolsep}{3.2pt}
\renewcommand{\arraystretch}{1.08}

\begin{tabular}{lccc}
\toprule
Configuration
& $\mathrm{MAE}_{\mathrm{int}}$
& \makecell{Exact Match\\(\%)}
& $R^2_{\mathrm{int}}$ \\
\midrule

Proposed
& 0.0815 $\pm$ 0.0018
& 57.75 $\pm$ 0.85
& 0.7769 $\pm$ 0.0111 \\

$224 \times 224$ resolution
& 0.0826 $\pm$ 0.0027
& 58.34 $\pm$ 1.29
& 0.7734 $\pm$ 0.0031 \\

Bilinear interpolation
& 0.0817 $\pm$ 0.0017
& 57.72 $\pm$ 0.28
& 0.7755 $\pm$ 0.0071 \\

Channel repetition
& 0.0823 $\pm$ 0.0029
& 57.99 $\pm$ 1.30
& 0.7731 $\pm$ 0.0054 \\

Without temporal warping
& 0.0900 $\pm$ 0.0018
& 55.62 $\pm$ 1.17
& 0.7488 $\pm$ 0.0049 \\

Feature-axis permutation
& 0.0898 $\pm$ 0.0024
& 56.98 $\pm$ 1.18
& 0.7522 $\pm$ 0.0045 \\

Temporal-axis permutation
& 0.2035 $\pm$ 0.0020 
& 31.18 $\pm$ 0.80 
& 0.3721 $\pm$ 0.0084 \\

From scratch
& 0.2849 $\pm$ 0.0002
& 4.99 $\pm$ 0.50
& $-0.2002 \pm 0.0099$ \\

\bottomrule
\end{tabular}
\end{table}

\subsection{Feature-Space Characterization}
\label{sec:feature}

\begin{figure}[t]
\centering
\includegraphics[height=4cm, keepaspectratio]{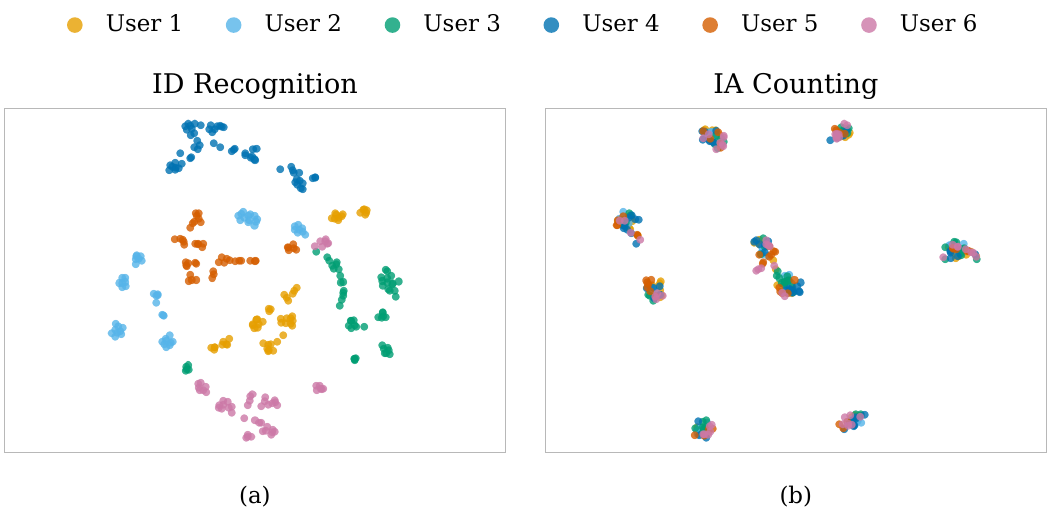}
\caption{Separately fitted t-SNE projections of identity-dependent and
identity-agnostic embeddings, colored by user identity (5~GHz, all
environments, Standard protocol, data partition 1, single-user samples).}
\label{fig:tsne}
\end{figure}

Cosine-distance silhouette scores (Table~\ref{tab:feature_space}) indicate
that the identity-dependent embeddings were organized more strongly by user
identity than by activity composition (0.345 versus 0.244), whereas the
identity-agnostic embeddings showed the reverse ($-0.037$ versus 0.669). The
negative identity-agnostic user score indicates that samples lay on average
closer to other users' samples than to their own. The t-SNE projections
qualitatively support this pattern (Fig.~\ref{fig:tsne}): users occupy
largely separate regions under identity-dependent recognition but are
substantially intermixed under identity-agnostic counting.

The linear count probe separates the two contributions. Under the Standard
protocol the probe on identity-agnostic features attained
$\mathrm{MAE}_{\mathrm{int}} = 0.0830 \pm 0.0022$ and
$R^2_{\mathrm{int}} = 0.7733 \pm 0.0084$, against $0.1121 \pm 0.0022$ and
$0.6932 \pm 0.0195$ on identity-dependent features. The identity-agnostic
probe recovered almost all of the end-to-end performance (0.0815), so the
linear readout was not the limiting factor. Under LUO the gap persisted,
$0.0865 \pm 0.0171$ against $0.1120 \pm 0.0146$, with the identity-dependent 
probe reaching a low and highly variable $R^2_{\mathrm{int}}$ of $0.0830 \pm 0.1611$.

\subsection{Ablation Study}
\label{sec:ablation}

Spatial resolution, interpolation method, and channel projection strategy
left performance essentially unchanged, with $\mathrm{MAE}_{\mathrm{int}}$
differences of at most 0.0011, small relative to the observed
across-partition variation (Table~\ref{tab:ablation}). Removing temporal
warping increased $\mathrm{MAE}_{\mathrm{int}}$ from 0.0815 to 0.0900.

Two further variants permuted one axis of the resized CSI representation
using the same fixed, one-to-one mapping for all training and test samples.
This retained all values but rearranged their spatial neighborhoods. Permuting the antenna--subcarrier axis
increased $\mathrm{MAE}_{\mathrm{int}}$ to 0.0898, a change comparable to
removing temporal warping, whereas permuting the temporal axis increased it
to 0.2035 and reduced exact-match accuracy to 31.18\%, while remaining well
away from the all-zero solution on every metric. Training from scratch
increased $\mathrm{MAE}_{\mathrm{int}}$ by a factor of 3.5 to 0.2849 and
yielded an $R^2_{\mathrm{int}}$ of $-0.2002$; as in the identity-agnostic
implementation of THAT (Section~\ref{sec:backbone}), this configuration
assigned zero to every activity count.

\section{Discussion}
\label{sec:discussion}

\subsection{Determinants of Pipeline Performance}

The ablations reveal three levels of sensitivity. Spatial resolution,
interpolation, and channel strategy changed $\mathrm{MAE}_{\mathrm{int}}$ by
at most 0.0011. Removing temporal warping or permuting the
antenna--subcarrier axis had moderate effects, whereas temporal permutation
and training from scratch were most damaging. The pipeline is therefore
robust to minor projection choices but relies on temporal locality and
pretrained initialization. Temporal permutation still outperformed the
all-zero predictor, indicating that useful scene information remained despite
the disrupted locality. By contrast, training from scratch and the
identity-agnostic implementation of THAT produced identical count metrics and
all-zero rounded outputs ($\mathrm{MAE}_{\mathrm{int}}=0.2849$), suggesting
that zero-inflated targets can induce the same degenerate solution across
models of different capacities under the evaluated optimization setting.

Combined-environment training outperformed environment-specific training,
but this may reflect both its roughly threefold larger training set and
exposure to diverse multipath structures. The consistent 5~GHz advantage 
is expected: the wider channel bandwidth gives finer frequency resolution 
and the shorter wavelength increases sensitivity to small-scale body displacement.

\subsection{Why Counting Outperforms Identity-Indexed Prediction}

The two formulations share the same preprocessing and backbone and differ only
in output head and objective, isolating the effect of problem formulation.
Despite receiving coarser supervision, identity-agnostic counting is more
accurate on every count metric except exact match, and this ordering holds in
each of the eight environment--band conditions.
The separation widens sharply under user shift. There the derived counts are
barely more accurate than predicting zero for every activity, while
identity-agnostic counting attains roughly half that error. This difference is
not visible under closed-set evaluation.

The training objectives provide a plausible explanation. User-indexed
prediction must determine who performs each activity, encouraging features
linked to body-induced multipath signatures and habitual motion patterns.
These features can aid enrolled-user recognition but become unreliable under
user shift, and prioritising them may divert representational capacity from
activity-composition structure. Counting instead assigns the same target to a given activity composition
regardless of which users perform the activities. Its loss therefore does not
explicitly reward representations that separate users.

The representation analyses support this interpretation. Identity-dependent
embeddings cluster primarily by user identity, whereas identity-agnostic
embeddings cluster by activity composition and show substantial user mixing.
With identical frozen linear counting heads, identity-agnostic features
yield 26\% lower counting error than identity-dependent features
in-distribution and 23\% lower under user shift.
Because both probes used the same readout and objective, the remaining gap
cannot be attributed to the final prediction head; it reflects differences
already encoded in the representations by the original training
formulations. Environment shift remains challenging because
room-dependent multipath changes also affect activity-related components; as
fractions of the zero-predictor error the gap is smaller under LOEO (0.66
against 0.88) than under LUO (0.51 against 0.96). The widening gap at five
users is consistent with the additional slot-to-user assignment burden of
identity-indexed prediction.

\subsection{Metric Behaviour and Error Sources}
\label{sec:metrics_behaviour}

The higher closed-set exact-match accuracy of the derived counts does not
imply more accurate counting, since their cell-wise accuracy is lower and
their MAE is higher. The error distribution reported in
Section~\ref{sec:closedset} explains the difference: derived count errors are
concentrated in fewer samples but affect more entries within them.
Substituting one activity for another changes two count-vector entries,
whereas an activity--$\emptyset$ confusion changes one; conversely, permuting
activities between user slots penalises identity recognition while leaving
the aggregate count vector unchanged. These mechanisms explain how the
derived counts can lead in closed-set exact-match accuracy despite worse
cell-wise counting performance.

Rounding generally reduces MAE because most continuous predictions lie
closer to the correct integer, but it can lower $R^2$ when predictions cross
integer boundaries. We therefore treat $\mathrm{MAE}_{\mathrm{int}}$ as the
primary metric because it evaluates both formulations in the same integer
count space, is directly interpretable as the mean count error per
activity--sample entry, and is not normalized by protocol-specific target
variance.

Both formulations struggle with similar activities, indicating a shared
component of signal ambiguity without excluding formulation-specific
effects. \textit{Sit down} and \textit{stand up} involve similar vertical
displacements distinguished mainly by temporal order, while distinguishing
\textit{Nothing} from $\emptyset$ depends on static multipath cues separating
a stationary user from an absent one. \textit{Nothing}'s confusion with
\textit{walk} and \textit{stand up} is consistent with these activities
ending in a stationary posture. \textit{Rotation}, \textit{walk}, and
\textit{jump} also share localised motion signatures. Error increases with
occupancy as concurrent movements produce overlapping channel effects,
while the near-zero error for empty scenes indicates that the absence of
activity is reliably detected.

\subsection{Practical Implications and Limitations}
\label{sec:implications}

The identity-dependent formulation ties output slots to physical
individuals, requiring prior enrolment and retraining when the population
changes. Identity-agnostic counting instead produces a fixed-length vector
determined only by the activity set and can operate without a predefined
user roster. Its output dimensionality is independent of the maximum
supported occupancy, although reliable count ranges remain bounded by the
occupancies observed during training. Summing the predicted activity counts
also provides an implicit occupancy estimate without a separate prediction
head. For deployment, parameter and MAC reductions did not translate
proportionally into CPU latency, so hardware-specific runtime and memory
measurements are more informative than complexity proxies alone.

Because its output describes the scene rather than specific individuals,
counting avoids explicit identity linkage at the point of reporting: a count
vector specifies how many users perform each activity without identifying
those users. This output-level data minimisation is not a formal privacy
guarantee, since the underlying CSI and learned representations may still
contain recoverable user-specific information. Counting therefore complements
rather than replaces signal- or representation-level privacy mechanisms. It
is well suited to aggregate activity and occupancy monitoring, but cannot by
itself provide the event-to-person attribution required in personalised
healthcare, assisted living, access control, and individual-level safety
monitoring.

The evaluation is confined to WiMANS, which, to our knowledge, remains the
only publicly available dataset for multi-user WiFi activity sensing. Its
recordings therefore share a single hardware platform and capture
configuration, and the flattened representation used here is tied to that
configuration's antenna and subcarrier counts. Within these limits the
lower-MAE ordering is consistent, repeating across all eight
environment--band conditions and the Standard, LUO, and LOEO protocols, but
whether it transfers to other deployments remains open. Two boundaries of the
setting are worth noting. Scenes contain at most five concurrent users, and
the gap between formulations widens at the highest observed occupancy, so
denser scenes are the natural next test rather than a safe extrapolation.
Only CSI amplitude is used, since raw phase from commodity devices carries
hardware-induced offsets that require sanitisation; recovering it may improve
the separation of activities with similar amplitude signatures, such as
\textit{sit down} and \textit{stand up}.

\section{Conclusion}

This work compared identity-dependent recognition with identity-agnostic
activity counting using matched integer-valued count metrics and a shared
pipeline that isolates the effect of output structure and objective. Despite
using coarser supervision, identity-agnostic counting achieved lower MAE,
higher cell-wise accuracy, and higher explained variance under closed-set
conditions, repeating across all eight environment--band conditions; derived
counts led only in exact-match accuracy, reflecting a different concentration
of errors rather than lower overall count error. The two formulations
separated further under domain shift, where identity-agnostic counting led on
every counting metric under both unseen users and unseen environments, and
the derived counts remained close to the all-zero predictor under unseen
users. Closed-set evaluation therefore overstates how well identity-indexed
prediction recovers scene-level activity composition. Representation analyses
indicate why: identity-dependent embeddings were organized primarily by user
identity and identity-agnostic embeddings by activity composition, and a
linear probe on frozen features localized the difference to the
representation rather than the output head. Requiring a model to decide who
performs each activity is therefore not a neutral design choice, since it
commits capacity to structure that is informative only for enrolled users.
Where the question is what is happening in a scene rather than who is doing
it, activity counting is the more accurate and robust formulation, and
operates without a predefined user roster; extending it to denser scenes and
other capture configurations is the natural next step.

\section*{Acknowledgment}
This work was supported by the Engineering and Physical Sciences Research Council [grant number EP/Z00084X/1]. For the purpose of open access, 
the author has applied a Creative Commons Attribution (CC BY) licence to any Author Accepted Manuscript version arising.

\bibliographystyle{IEEEtran}
\bibliography{references}

\vfill

\end{document}